# The Evolution of Learning Algorithms for Artificial Neural Networks*

Jonathan Baxter
*School of Information Science and Technology*
*The Flinders University of South Australia*
Email: `jon@maths.flinders.edu.au`

**Abstract.**
   In this paper we investigate a neural network model in which weights between computational nodes are modified according to a *local learning rule*. To determine whether local learning rules are sufficient for learning, we encode the network architectures and learning dynamics genetically and then apply selection pressure to evolve networks capable of learning the four boolean functions of one variable. The successful networks are analysed and we show how learning behaviour *emerges* as a distributed property of the entire network. Finally the utility of genetic algorithms as a tool of discovery is discussed.

## 1. Introduction

Artificial neural networks (henceforth ANNs) are biologically motivated computational models in which large numbers of nodes communicate with one another via weighted connections (a weighted connection is more commonly called a *weight*). The operation of nodes is abstracted from the behaviour of real neurons; the *activation level* of each node is determined by the activation levels of nodes connected to it, and the relative strengths of the connections (weights), in the same way that the firing rate of a real neuron is determined by the firing rates of neurons connected to it, and the efficacy of their synaptic connections. The weights in an ANN are a simple means of modelling synaptic strengths, an excitatory synapse is represented by a positive weight, while an inhibitory synapse is represented by a negative weight. Currently the most popular kind of ANN is the *backpropagation network* [1] in which the nodes are partitioned into separate layers and each node is connected by weights only to those nodes in the layer above. Referring to Figure 1, backpropagation networks are trained to classify patterns appearing on their inputs into various output classes by dynamic adjustment of the weights.

   Although there is increasing evidence that animal learning is facilitated by changes in synaptic strengths, it is unlikely that animal networks employ any of the mechanisms used in training backpropagation networks (gradient descent, stochastic descent, genetic algorithms, etc.) because they all use information from across the entire network to decide how to adjust the value of an individual weight. Such information is unlikely to be available to a synapse in a biological network; the only information it can possess is of a local origin, primarily

---

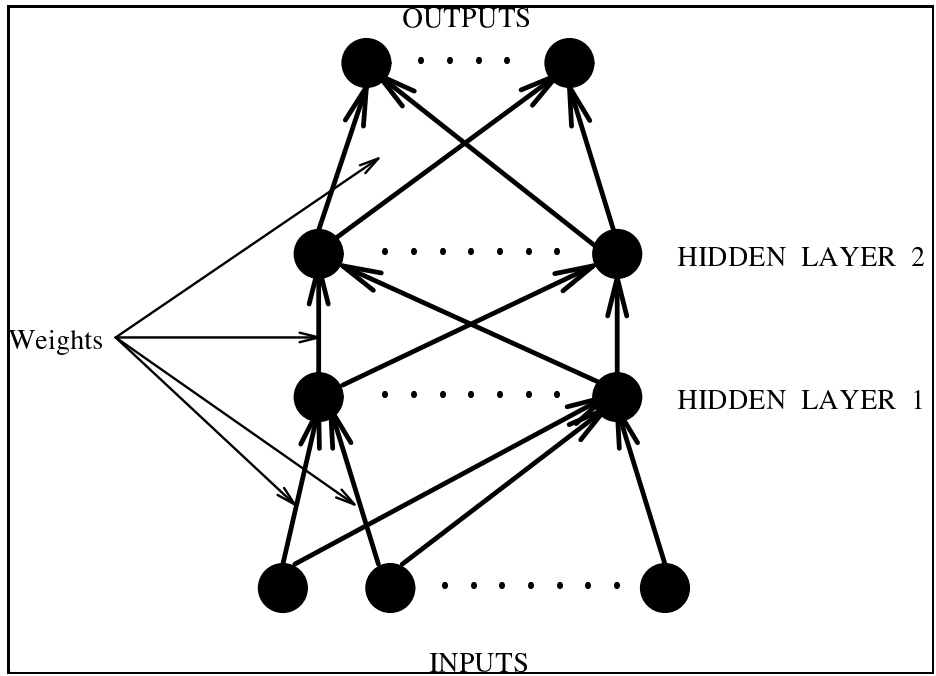

Figure 1: Backpropagation network.

presynaptic and postsynaptic potentials. Thus a biologically plausible training algorithm for ANNs would use only local information when adjusting weights. What then is the information local to a weight in an ANN? Looking again at Figure 1 we see that a weight is really only in direct communication with the node it emanates from, and the node it feeds into, so we make the following anzatz:

> *The information local to a weight in an ANN is the activation levels of the two nodes it connects.*

A natural question now arises, "How does learning occur if weights only change in response to their local environment?"

This paper answers the question for an extremely simple neural network model in which neurons have only two activation levels, $\pm 1$, weights only come in three flavours — $+1$ for excitatory connections, $-1$ for inhibitory connections, and zero for no connection at all — and when the weights change, if at all, they do so only in response to changes in the activation levels of the nodes they connect. We call such networks *local binary neural nets* (LBNNs).

Designing LBNNs to solve particular learning tasks is made difficult by the complex behaviour of even simple networks of this type, and at the outset it is not clear that any LBNN will exhibit useful learning behaviour. However, animal brains with sophisticated learning abilities evolved by a process of replication and natural selection, so if LBNNs capture the essential features of animal computation then the simulated evolution of LBNNs might produce one with learning ability. Thus, rather than designing an LBNN, we tried evolving one.

The application of evolutionary techniques to solving problems usually goes by the name *genetic algorithm* [2]. Genetic algorithms are not new in neural networks research, but to date they have primarily been used as a learning algorithm, that is, as a method for setting the weights in feedforward and recurrent nets (see e.g. [3, 4]), or as a method for generating initial network topologies to which traditional learning techniques can be applied (e.g. [5]). The logical precursor to the work presented here is [6] in which a genetic algorithm was used to evolve the well-known delta rule for single layer feedforward networks. It is the use of a GA to search for learning algorithms themselves, rather than as a learning algorithm itself, that distinguishes both [6] and the work presented here. One important difference between the work in [6] and our work is that we have used the genetic algorithm to search in a relatively uncharted area of network learning, an area with no known solutions.

## 2. Local Binary Neural Nets (LBNNs)

LBNNs consist of interconnected binary computational units (nodes) operating in a discrete-time, synchronous fashion. The $n$ nodes in our network are divided into the three classes (input, hidden and output) already mentioned, however the architecture, i.e. which nodes are connected, is completely unrestricted, rather than layered as in backpropagation networks. The operation of the network is divided into two phases, *training* and *testing*. During the training phase, the input and output nodes are clamped by the environment, whereas during the testing phase only the input nodes are clamped. The idea is to teach the network to associate certain outputs with certain input vectors during the training phase, and then to recall the learned outputs upon reapplication of the input vector during the testing phase. More details of the training and testing of the network are given in the next two sections, after we outline the network dynamics below.

Denoting the activation of any unclamped node $i$ at time $t$ by $a_i(t)$ and the connection strength between nodes $i$ and $j$ by $w_{ij}(t)$, the activation of node $i$ at time $t+1$ is computed by:

$$a_i(t+1) = \theta \left( \sum_{j=1}^{n} a_j(t) \cdot w_{ji}(t) \right) \quad (1)$$

where $\theta$ is the sign function:

$$\theta(x) = \begin{cases} 1 & \text{if } x > 0 \\ -1 & \text{if } x \leq 0 \end{cases}$$

To simplify matters we only allow weight values of $+1$, $-1$ or $0$. A zero weight, $w_{ij} = 0$, between node $i$ and node $j$ means that node $i$ is not connected to node $j$. Weights in our networks come in two varieties, *fixed* and *learnable*. The value of a fixed weight does not change over time:

$$w_{ij}(t+1) = w_{ij}(t)$$

whereas the value of a learnable weight changes at each time step according to the *local learning rule $f$*:

$$w_{ij}(t+1) = f(a_i(t), a_j(t)), \quad (2)$$

where $f$ is time-independent and the same for all weights. Only non-zero weights can be learnable, so two unconnected nodes forever remain unconnected. Also, we do not allow a learnable weight to become zero, so the learning rule $f$ is a $\pm 1$ valued function. This learning rule is local in both space and time as the new weight value depends only upon the preceding activation levels of the two nodes it connects and upon no activation levels or weight values from anywhere else in the network.

We can consider (2) to be a generalization of the well-known *Hebbian Learning Rule*, postulated by Hebb [7] as a mechanism of learning at the cellular level in animal brains. Hebb suggested that synaptic strengths increase between neurons whose activation levels are correlated, and likewise decrease between neurons whose activation levels are uncorrelated. Translated into our current framework this means that for a learnable weight $w_{ij}$, $w_{ij}(t+1) = 1$ whenever $a_i(t) = a_j(t)$, and $w_{ij}(t+1) = -1$ whenever $a_i(t) \neq a_j(t)$. As all activation values are binary ($\pm 1$) this translates into the following formula for the Hebb learning rule:

$$f_{\text{hebb}}(a_i(t), a_j(t)) = a_i(t) \cdot a_j(t) \qquad (3)$$

### 2.1. Training the network

As our networks are binary we evolve them to learn boolean functions with multidimensional boolean domains, that is, $\{\pm 1\}$ valued functions with domain $\{\pm 1\} \times \{\pm 1\} \times \ldots \times \{\pm 1\} \stackrel{\text{def}}{=} \{\pm 1\}^m$. Hence our networks have $m$ input nodes, one output node and $n - m - 1$ hidden nodes where $n$ is the total number of nodes in the network. A network is considered to have learnt the boolean function $g : \{\pm 1\}^m \longrightarrow \{\pm 1\}$ if any boolean vector $x \in \{\pm 1\}^m$ applied to its inputs causes the output node to converge to $g(x)$ after a suitable number of updates of the hidden and output nodes according to (1) and the learnable weights according to (2). We teach the network the input-output pair $\{x, g(x)\}$ by clamping its input node activations to the components of $x$, its output node activation to $g(x)$ and updating the hidden node activations and learnable weight values according to (1) and (2) a set number of times. The entire function $g$ is taught to the network by sequentially teaching it $\{x, g(x)\}$ for every $x \in \{\pm 1\}^m$.

Note that network training proceeds by simply showing the network what it should do for every input vector, a form of totally supervised training. Learning all takes place through the application of (1) and (2); we do not reach into the network and adjust hidden node activation levels, nor do we play around with the values of the learnable weights in order to achieve the desired network behaviour. We do not even indicate to the network that training is complete, we merely cease clamping the output node. With all these restrictions a network's learning ability must primarily be a property of its architecture, not of some sophisticated method for setting its weights (contrast this to backpropagation networks which have a complex algorithm to set the weights, but limited architectures).

### 3. Evolution of neural networks

The four important components of natural evolution for which we need to find artificial analogues are *genotype*, *phenotype*, *morphogenesis* and *fitness criterion*. All living organisms use DNA as their genotype, which can be thought of as a (very long) base-four number, so in our simulations we use bit-strings, i.e. base-two numbers, for the genotype. The choice for phenotype is clear — Local Binary Neural Networks. Morphogenesis is the process of turning

DNA into a viable phenotype, so we must decide how to turn bit-strings into LBNNs. This is discussed in the next section. Finally, the fitness criterion is a neural network's ability to learn a set of boolean functions, the greater its learning ability the higher its fitness. We discuss the details of the fitness criterion later, after we have outlined the evolutionary procedure.

Evolving a population of neural networks proceeds as follows. Initially generate a random population of bit strings (genomes), and decode the strings (morphogenesis) into their associated networks (phenotypes). Determine the fitness of each network in the initial population, which in our case means determining each network's learning ability. Next generate a new population by selecting pairs of strings from the initial population (with replacement) in a probabilistic fashion based upon the observed fitness of their associated networks (strings corresponding to fitter networks are given a higher probability of being selected than strings with lower fitness). Mate the strings together to produce two offspring by cutting them both at some random point along their length (the same cutpoint is used on both strings) and swapping over the genetic material after the cutpoint. This process is commonly known as *crossover*. Each bit in the new strings also has a 1% chance of mutating,[1] i.e. flipping from 0 to 1 or vice-versa. Once enough strings have been generated to form a new population, the old population is discarded (except for the best few strings) and the process of fitness determination is repeated on the new population. The process of fitness determination, selection, crossover and mutation is repeated for many generations, either until a network is found with the desired fitness level, or we give up.

We now outline how the networks are encoded as bit strings, and the details of the fitness criterion.

*3.1. Representation scheme*

The representation of neural networks as bit strings used here is very straightforward. It is assumed that each network in the population has the same number of nodes, $n$, which is fixed for the whole evolutionary run. To completely specify a network we must specify both its architecture and the learning rule employed. From (2) we see that the learning rule is simply a boolean function of two variables and hence requires four bits for its specification, one bit for each of the possible values of the vector $(a_i(t), a_j(t))$. The network architecture is specified by determining which nodes are connected by non-zero weights, whether those weights are learnable or fixed, and for the fixed weights, their values. We code this information using three bits for each pair of nodes in the network. The first bit indicates whether the weight is non-zero, the second bit whether the weight is learnable or fixed, and the third bit the value of the weight. Obviously this is an inefficient coding (for instance the third bit will only be significant if the weight is fixed, as learnable weights start out with random values), but it allows us to have uniform length bit strings, hence avoiding complications involved in crossing over strings of different lengths. Networks of $n$ nodes are thus represented by bit strings of length $3n^2 + 4$, although because of the inbuilt redundancy of our representation scheme the mapping between strings and neural networks is not one-to-one.

---

[1] A mutation rate of 1% was chosen fairly arbitrarily, somewhere between being too disruptive (i.e. too large) and too small to have significant effect.

Table 1: Boolean functions of one variable.

| Function | Input | Output |
|---|---|---|
| $f_0$ | $-1$ | $-1$ |
|  | $1$ | $-1$ |
| $f_1$ | $-1$ | $-1$ |
|  | $1$ | $1$ |
| $f_2$ | $-1$ | $1$ |
|  | $1$ | $-1$ |
| $f_3$ | $-1$ | $1$ |
|  | $1$ | $1$ |

*3.2. The fitness criterion*

The fitness criterion we chose was a network's ability to learn the four boolean functions of one variable (see Table 1). Note the boolean functions have domain and range $\{\pm 1\}$ rather than $\{0, 1\}$, but this does not alter their boolean character. Thus our networks have one input and one output node in addition to an as yet unspecified number of hidden nodes. We chose such a simple learning task because it is easily shown using an inductive argument (on the number of input variables) that networks capable of learning the boolean functions of one variable can be wired together to form networks capable of learning all boolean functions of an arbitrary number of variables. A network was tested for its learning ability by training it on each of the four functions according to the method outlined in Section 2.1. Before training on each function, the network's hidden nodes and learnable weights were randomly initialised (to $\pm 1$). We used four learning cycles during the training phase,[2] that is, the hidden node activations and learnable weight values were updated four times after the application of each training pair. During testing the networks were allowed four cycles with the input node clamped for the output node to converge, and then the output node value was recorded for a further four cycles to measure its stability. A score was kept of the number of errors made by the network, the maximum possible being 32 (four for each of the two training pairs in each of the four functions). Denoting the score by $N$, the fitness of a network was given by:

$$\text{fitness} = \frac{1}{1 + N}$$

A network with no errors received a fitness of 1, whilst a completely confused network gained a fitness of $\frac{1}{33}$.

*3.3. Function and training pair ordering*

In our initial experiments we presented the training pairs and the functions in the same order for the whole evolutionary run. We also cycled the networks the same number of times every time a training pair was presented. This led to networks that were highly sensitive to the presentation order of the functions, the order of the training pairs within each function, and to the number of network updates used during learning. The networks would only learn correctly under conditions identical to those in which they had evolved, but could not perform when conditions were varied even slightly. This undesirable behaviour was avoided in later experiments by testing the networks more than once per generation, each time with a randomly generated environment. In particular, the functions were presented in a random order, the training pairs for each function were presented in a random order, and the number of learning cycles was randomly chosen in the range 4–8. In this way a network could not achieve a high fitness by being specifically adapted to one particular environment; it had to perform well under several different sets of conditions, and moreover the conditions altered from generation to generation. We experimented with the number of tests required, from two to six per generation, and found that five was the smallest number to consistently produce networks with general learning behaviour.

---

[2] "Four" was determined experimentally; evolutionary runs with fewer learning cycles did not yield any functional networks.

One may ask why this strategy did not simply produce networks specifically adapted to five distinct environments, but unable to perform well under any other circumstances? There are two main reasons why this did not happen. First, even if a network was specifically adapted to the five environments in its generation, its offspring were unlikely to be highly fit in the next generation because the parent's adaptations would no longer be of any use in the new environment. So in the long run, genetic material conferring an advantage on networks only in specific environments did not compete with material that conferred an advantage under a wider variety of conditions. Second, we would expect that at some stage it would be "easier" for a network to employ a learning strategy that worked independently of its environment, rather than employing a strategy only suited to a particular set of environments. For our nets it appears that learning in five environments is about as hard as learning in all environments, and so a network that could successfully learn the functions under five different circumstances was fairly likely to learn under all circumstances.

## 4. Results

All simulations were run on the ANU connection machine (CM2). The connection machine is a massively parallel SIMD machine consisting of 16384 bit-serial processors, usually partitioned into two groups of 8192 each. Tailoring our algorithm to the machine architecture, we ran all simulations with a population size of 8192. This enabled us to simulate the fitness calculations of an entire generation in parallel, one network to each processor. The relatively large population size also enabled faster convergence to optimal solutions.

We searched for networks that could learn with the minimum number of hidden nodes, and so initially we tried evolutionary runs using networks with zero, one and two hidden nodes. None of these simulations converged — we gave up after no improvement in the maximum fitness had occurred for fifty generations. Three hidden nodes seems to be the minimum required for this task, and we ran five simulations in all with this number, each starting with a different random number seed. In all runs the best network in each generation was retested under every possible environment, with 16 different random initialisations of the hidden nodes and learnable weights, and its performance averaged to produce a mea-sure of its environment-independent fitness[3]. Four of the five runs produced networks with an environment-independent fitness of 1 (the maximum possible, indicating perfect learning under all conditions), the quickest doing so in 35 generations. The fifth run failed to produce a network with best fitness in 200 generations, at which point we gave up. (Incidentally, 200 generations takes about 90 minutes of cpu time on the connection machine.) The best network's fitness, its environment independent fitness, and the average fitness across the whole population are plotted in Figure 2 for a typical evolutionary run. Notice that the environment-independent fitness stays low for several generations after the maximum fitness reaches 1, indicating that the best networks at this stage are not able to learn in all environments.

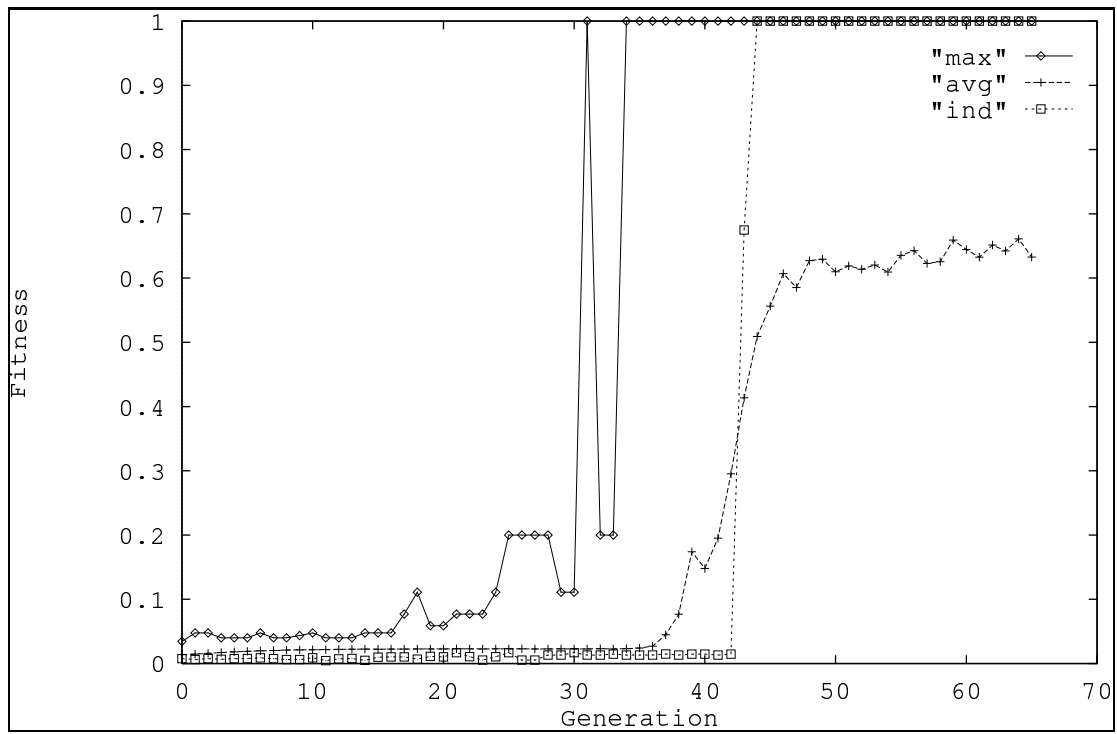

Figure 2: Maximum, average and environment-independent fitness over a typical evolutionary run.

### 4.1. Behaviour of the successful networks

Perhaps the most surprising result is that in all five simulations Hebb's rule (3) evolved as the local learning rule. Given that we were searching for networks with the smallest number of nodes this leads us to conjecture that a network employing Hebb's rule needs the fewest number of hidden nodes in order to learn the boolean functions of one variable. If Hebb's rule is indeed the most efficient, this would go some way towards explaining why it is the one employed in animal brains.

---

[3]There were 7680 environments in the environment independent test — 16 random hidden node and learnable weight initialisations times 24 orderings of the four functions times 2 orderings of the training pairs during training times 2 orderings of the training pairs during testing times 5 for the number of different learning cycle values.

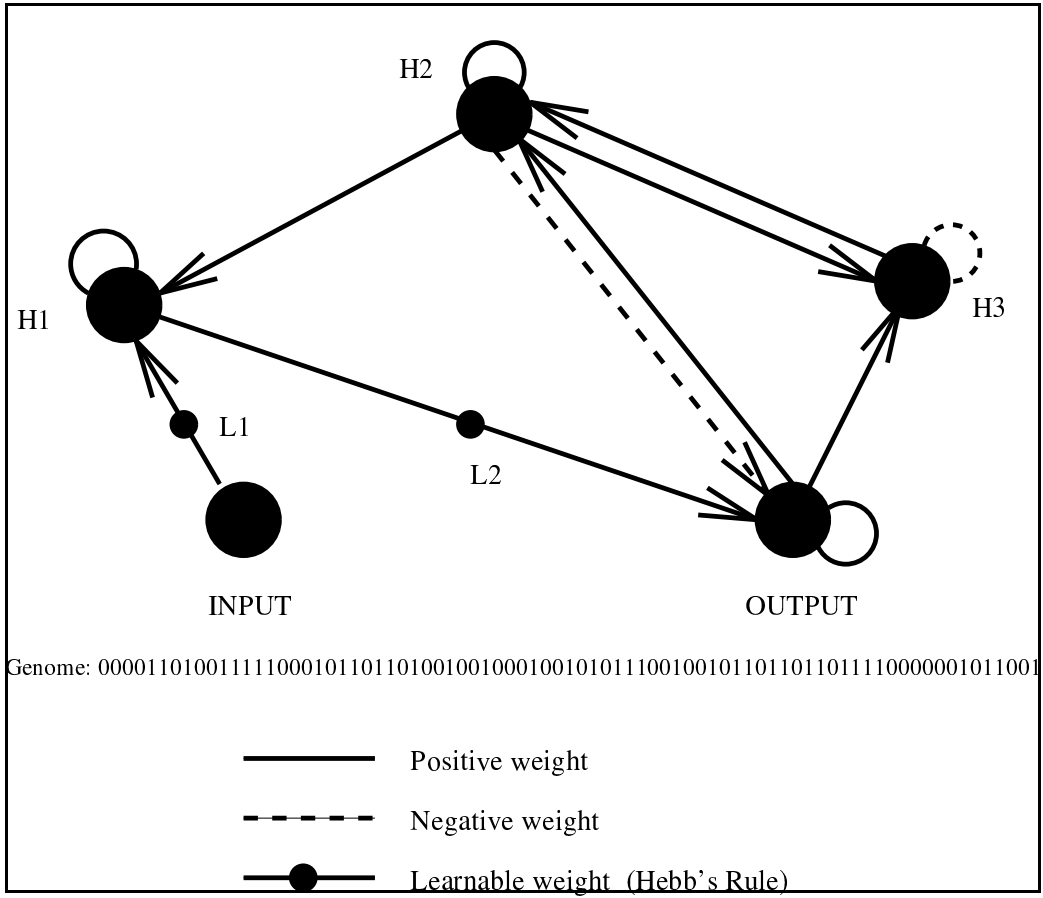

Figure 3: LBNN capable of learning all four boolean functions of one variable.

A typical successful network is shown in Figure 3. Note that we have not drawn any connections going into the input node, as the input node is always clamped by the environment and hence such connections have no effect on its activation level. In order to simplify the analysis, an extra evolutionary run was performed in which the networks were penalised slightly for having large numbers of learnable weights, hence the successful network shown in Figure 3 has been explicitly selected to have the minimum number of learnable weights.

The best way to understand how the network learns is to think of it as a discrete-time, discrete state-space dynamical system. At any point in time the state of a network is uniquely determined by its node activation levels and the values of its learnable weights, with its state at the next time point then given by (1) and (2). The network in Figure 3 has two learnable weights and five nodes so its state can be represented by a seven-bit vector. Thus the state-space of this network is the set of all length seven binary numbers. Given any initial state, $x_0$, the network will follow a path through state space, $x_0 \to x_1 \to x_2 \to \ldots$, determined by (1) and (2). A state which is mapped to itself under the action of (1) and (2) is called a *fixed point* of the network ($x \to x$). A state that maps back to itself in $m$ time steps and no fewer is called a *fixed point of period $m$* ($x_0 \to x_1 \to \ldots \to x_{m-1} \to x_0$). We generically denote fixed points and fixed points of period $m$ by the term *periodic points*. Note that the state space has 128 points so the maximum possible period is 128. The *basin of attraction*, $B(x)$, of a periodic point $x$ is the set of all points that eventually map to $x$ under the action of (1) and (2). Every point in the network's state space is in the basin of attraction of a unique cycle of periodic points, hence the network's behaviour is determined by its periodic points and their associated basins of attraction.

As the input node is always clamped by the environment, the network's behaviour divides into two categories — what it does with an input of $+1$, and what it does with an input of $-1$. It turns out that this net has four distinct fixed points of period one for each of the input values, and no higher period fixed points. Table 2 lists the fixed points with clamped inputs, and their basins of attraction. Note that for clarity we have written "0" instead of "$-1$". The same fixed points with the input clamped are also fixed points of the network when both the input and output are clamped, and moreover these are the only periodic points of the network with both input and output clamped. Hence, regardless of how we set the input and output nodes of the network, if we cycle the network for long enough it will end up in one of the fixed points listed in Table 2, and so in order to understand the network's behaviour we just need to understand how it moves between the fixed points in response to changes in its input and output node activations.

Let $(1, x_0)$, $(1, x_1)$, $(1, x_2)$, $(1, x_3)$ be the four fixed points with input 1, and $(-1, x_0')$, $(-1, x_1')$, $(-1, x_2')$, $(-1, x_3')$ be the four fixed points with input $-1$. Notice from Table 2 that $(1, x_i') \in B(1, x_i)$ and $(-1, x_i) \in B(-1, x_i')$ for $0 \leq i \leq 3$ ($(1, x_i')$ and $(-1, x_i)$ have been highlighted in Table 2, remembering that $-1$ is written as 0 for clarity). Thus $(1, x_i)$ is paired with $(-1, x_i')$ in the sense that if the network is in state $(1, x_i)$ and we set the input to $-1$, then after a number of updates the state of the network will converge to $(-1, x_i')$ (and vice-versa). Also, each pair $\{(1, x_i), (-1, x_i')\}$ represents a distinct boolean function, as we can see by writing the output node value out explicitly (see Table 3). Thus, for the network to learn function $f_i$, it must be induced into either of the states $(1, x_i)$ or $(-1, x_i')$. The fixed point structure of the network is such that this is achieved by clamping the input-output nodes to each of the pairs $(1, f_i(1))$, $(-1, f_i(-1))$, as is required for the network to have perfect fitness, and we illustrate this in a state transition diagram for the network, each state being a fixed point and the transitions caused by alteration of the input or output nodes (Figure 4).

## 4.2. Emergence of learning behaviour

It should be clear from the preceding section that the learning ability of the network is distributed across the entire network. The network does not naturally divide into subunits, each unit performing a certain task such as learning one of the two input-output patterns; rather learning ability is an emergent property of the dynamics of the network, and depends critically on all aspects of the network's construction.

Interestingly, the learnable weights are not used by the network to explicitly store information. Naively one might imagine that a network with two learnable weights would learn a boolean function of one variable, $f$, by storing $f(1)$ in one of the learnable weights, and $f(-1)$ in the other. However it is clear that no such thing is happening; the learnable weights aid in the forming of appropriate fixed point structures rather than explicitly storing function information.

Table 2: Fixed points and attracting basins of network in Figure 3.

| Fixed Point (InputOutputH1H2H3L1L2) | Attracting Basin | | |
|---|---|---|---|
| *0000100* $(0, x'_0)$ | 0101000 0000100 0101100 0110110 0011110 0011001 0011101 | 0011000 0010100 0011100 0001110 0111110 0111001 0110111 0011111 | *0111000* 0001100 0111100 0101110 0101001 0001101 0001111 |
| *0111010* $(0, x'_1)$ | 0000000 0110000 0100010 0101010 0100110 0110001 0110011 | 0100000 0001000 0010010 0111010 0010110 0001001 0000111 0010111 | 0010000 0000010 0110010 *0000110* 0100001 0100011 0100111 |
| *0111101* $(0, x'_2)$ | 0100100 0100101 0101101 | 0110100 0010101 0111101 0111111 | 0000101 0110101 0101111 |
| *0000011* $(0, x'_3)$ | 0001010 0010001 0001011 | 0011010 0000011 0101011 0111011 | *0000001* 0010011 0011011 |
| *1111000* $(1, x_0)$ | 1000000 1110000 *1000100* 1000010 1110010 1110001 1010101 | 1100000 1101000 1100100 1100010 1001010 1000101 1100011 1001011 | 1010000 1111000 1010100 1010010 1100001 1100101 1110011 |
| *1000110* $(1, x_1)$ | 1110100 1011100 1011010 1010110 1011110 1001101 1011011 | 1001100 1111100 *1111010* 1001110 1111110 1011101 1111011 1011111 | 1101100 1101010 1000110 1101110 1110101 1101011 1001111 |
| *1111111* $(1, x_2)$ | 1100110 *1111101* 1010111 | 1110110 1000111 1110111 1111111 | 1101101 1100111 1101111 |
| *1000001* $(1, x_3)$ | 1001000 1010001 1011001 | 1011000 1001001 1111001 1010011 | 1000001 1101001 *1000011* |

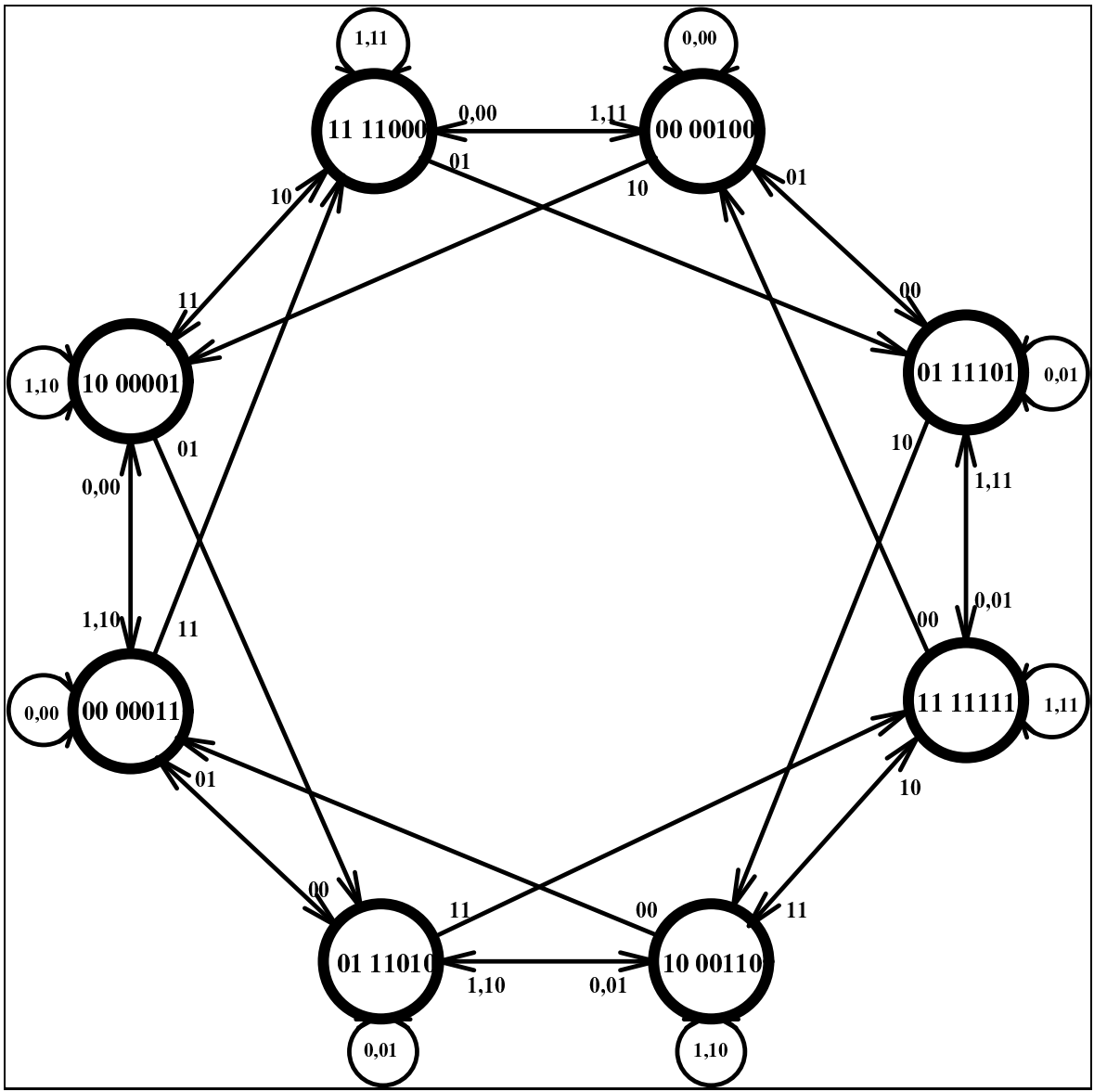

Figure 4: State transition diagram. Each node is labelled with a fixed point. Each vertical and horizontal edge of the octagon corresponds to one of the four boolean functions. The first two components of the fixed point are the values of the input and output nodes, the other components are the 3 hidden node values, and the 2 learnable weight values. The arrows indicate which fixed point the network will move to if the input and output nodes are set to the numbers at the root of the arrow. A number such as 1, 10 at the root of an arrow indicates that the network will move to the fixed point at the head of the arrow if its input node is set to 1 and its output is left free, or if its input is set to 1 and its output is set to 0. Arrows with just a pair of numbers at the root, (e.g. 01), indicate the network will move to the head of the arrow only if the input is set to the first of the pair (0), and the output is set to the second of the pair (1).

Table 3: Functions represented by fixed point pairs.

| Fixed Point Pair | Value (InputOutput H1H2H3L1L2) | Function $(1, f(1))$ $(0, f(0))$ |
|---|---|---|
| $(1, x_0)$ $(0, x_0')$ | (11 11000) (00 00100) | $(1, 1)$ $(0, 0)$ |
| $(1, x_1)$ $(0, x_1')$ | (10 00110) (01 11010) | $(1, 0)$ $(0, 1)$ |
| $(1, x_2)$ $(0, x_2')$ | (11 11111) (01 11101) | $(1, 1)$ $(0, 1)$ |
| $(1, x_3)$ $(0, x_3')$ | (10 00001) (00 00011) | $(1, 0)$ $(0, 0)$ |

## 5. Conclusion

In evolving neural networks with local learning rules we have taken two important hints from nature, one a plausibility argument for how learning algorithms should manifest themselves at a cellular level, the other the idea that complex systems with interesting behaviour can evolve as a result of continued selection pressure.

Although the networks are not directly useful (after all, who needs to learn the boolean functions of one variable?), they do provide important intuition about how learning behaviour can emerge from low-level computational structures, intuition that is invaluable when trying to design more practical devices, or when just theorizing about the role of computation in learning.

### Acknowledgements

Thanks to Shell Australia for providing me with a postgraduate scholarship, and also to my supervisor, Professor Bill Moran, for many helpful discussions during the course of this work.